%% file: main.tex
\documentclass[10pt,twocolumn,letterpaper]{article}

\usepackage{cvpr}      

\definecolor{cvprblue}{rgb}{0.21,0.49,0.74}
\usepackage[pagebackref,breaklinks,colorlinks,allcolors=cvprblue]{hyperref}

\def\paperID{9488} 
\def\confName{CVPR}
\def\confYear{2026}

\title{HulluEdit: Single-Pass Evidence-Consistent Subspace Editing for Mitigating Hallucinations in Large Vision-Language Models}

\author{
  Yangguang Lin$^{1}$, 
  Quan Fang$^{1}$$^\dag$, 
  Yufei Li$^{1}$, 
  Jiachen Sun$^{1}$, 
  Junyu Gao$^{2}$, 
  Jitao Sang$^{3}$ \\
  $^{1}$Beijing University of Posts and Telecommunications, Beijing, China \\
  $^{2}$Institute of Automation, Chinese Academy of Sciences, Beijing, China \\
  $^{3}$Beijing Jiaotong University, Beijing, China \\
  \texttt{\{sunshinelin, qfang, yufeili, sunjc\}@bupt.edu.cn} \\
  \texttt{junyu.gao@nlpr.ia.ac.cn} \\
  \texttt{jtsang@bjtu.edu.cn}
}
\usepackage{bm}
\usepackage{multirow}
\usepackage{multicol}
\usepackage{makecell}
\usepackage[table]{xcolor}
\usepackage{colortbl}
\usepackage{arydshln}
\begin{document}
\maketitle
\insert\footins{$\dag$ Corresponding author.}
\input{sec/0_abstract}
\input{sec/1_intro}
\input{sec/2_related_work}
\input{sec/3_method}
\input{sec/4_experiments}
\input{sec/5_conclusion}
\input{sec/6_acknowledgement}
{
    \small
    \bibliographystyle{ieeenat_fullname}
    \bibliography{main}
}

\clearpage
\appendix
\input{sec/appendix}

\end{document}

%% file: sec/0_abstract.tex
\begin{abstract}
    \textbf{Object hallucination} in Large Vision-Language Models (LVLMs) significantly hinders their reliable deployment. Existing methods struggle to balance efficiency and accuracy: they often require expensive reference models and multiple forward passes, or apply static edits that risk suppressing genuine visual evidence. To address this, we introduce \textbf{HulluEdit}, a \textbf{single-pass, reference-free} intervention framework. Our core innovation is \textbf{orthogonal subspace editing}: we decompose the hidden states of the model into orthogonal subspaces—visual evidence, conflicting priors, and residual uncertainty—enabling selective suppression of hallucinatory patterns without interfering with visual grounding. This approach mathematically guarantees that edits applied to the prior subspace leave the visual component entirely unaffected. Extensive experiments show that HulluEdit achieves state-of-the-art hallucination reduction on benchmarks including POPE and CHAIR across diverse architectures, while preserving general capabilities on MME and maintaining efficient inference. Our method consistently outperforms contrastive decoding and static subspace editing baselines, offering a new pathway toward more trustworthy LVLMs. Code is released at \url{https://github.com/VioAgnes/HulluEdit}.
\end{abstract}

%% file: sec/1_intro.tex
\begin{figure}[ht]
\centering
\begin{subfigure}{0.48\textwidth}
\includegraphics[width=\textwidth]{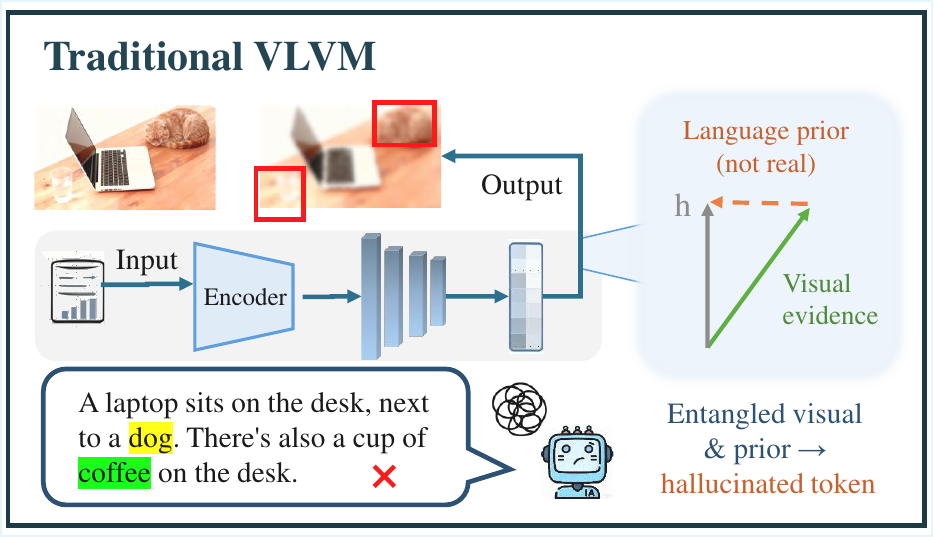}
\caption{Traditional LVLM with hallucinations}
\label{fig:traditional}
\end{subfigure}
\hfill
\begin{subfigure}{0.48\textwidth}
\includegraphics[width=\textwidth]{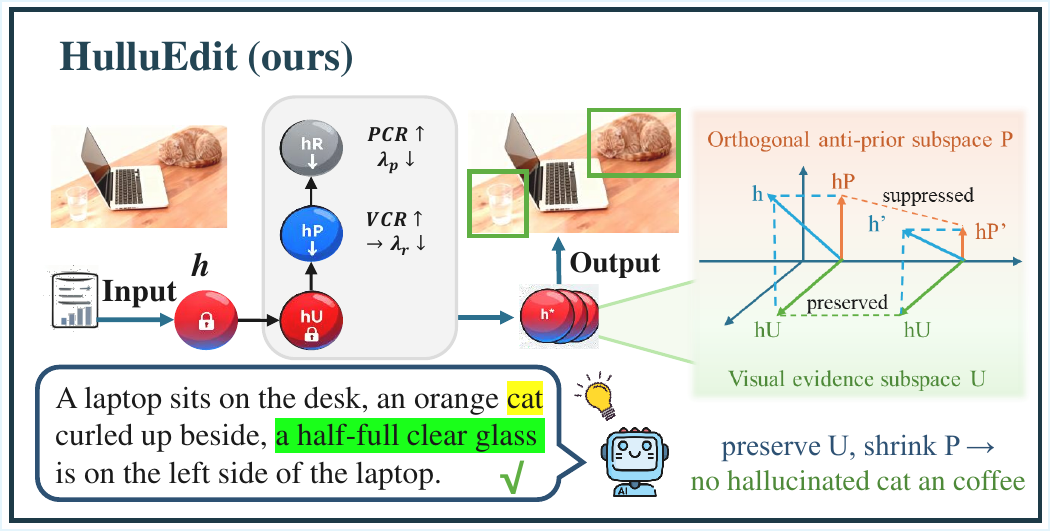}
\caption{HulluEdit (ours)}
\label{fig:hulluedit}
\end{subfigure}

\caption{Comparison of (a) traditional LVLMs prone to object hallucinations versus (b) our HulluEdit method that mitigates hallucinations via orthogonal subspace decomposition.}
\label{fig:framework_comparison}
\end{figure}

\section{Introduction}
\label{sec:intro}

Large Vision-Language Models (LVLMs) \cite{li2023blip2,zhu2023minigpt4,liu2023llava,alayrac2022flamingo,bai2023qwenvl} have become the standard foundation models for image captioning, visual question answering, and assistive interaction. However, they remain vulnerable to \emph{object hallucination} — the phenomenon of stating non-existent objects, attributes, or quantities when provided with an image \cite{rohrbach2018object,li2023evaluating,liu2024lvhallsurvey}. As illustrated in Figure~\ref{fig:traditional}, such hallucinations often arise when strong language priors override weak or ambiguous visual evidence, leading to a misalignment between the generated text and the actual image content.

To mitigate these hallucinations, researchers have proposed several strategies in recent years. \emph{Contrastive Decoding} methods \cite{li-etal-2023-contrastive,obrien2023cd,chuang2024dola,wang2024icd,leng2024vcd} achieve promising results by comparing output distributions, but often require a reference model or secondary inference, increasing latency and engineering complexity. Meanwhile, \emph{Static Subspace Editing} techniques \cite{yang2024nullu,Zhuang2025VASparse} construct dataset-level hallucination subspaces offline, but lack token-level adaptability and risk suppressing genuine visual evidence. Ultimately, as illustrated in Figure~\ref{fig:traditional}, these methods lack reliable decoupling mechanisms and fine-grained control between suppressing linguistic priors and preserving visual evidence.

Inspired by DeCo \cite{wang2025dcd}, we observe that representations from mid-layers of large models can serve as reliable references for calibrating the output layer. This motivates us to leverage intermediate layers for constructing sample-level subspace structures, enabling more refined and adaptive control over the evidence–prior competition. By further applying orthogonal decomposition to decouple priors from visual evidence, we can effectively suppress prior conflicts while preserving robust visual grounding.

Based on this insight, we propose \textbf{HulluEdit}, a single-pass subspace editing framework that decomposes hidden states into orthogonal components: an online-estimated \emph{Visual Evidence Subspace} and an \emph{Anti-Prior Subspace} in its orthogonal complement. As illustrated in Figure~\ref{fig:hulluedit}, we construct a sample-adaptive visual subspace via weighted SVD and characterize the anti-prior direction from a non-visual text cache. We then project hidden states onto these subspaces and perform component-wise editing, suppressing only the anti-prior components without interfering with visual evidence. Orthogonality guarantees that prior suppression does not damage visual grounding. The method requires no additional training, no secondary forward pass, and edits hidden states before the output layer logits, maintaining low overhead and ease of use.

Our main contributions are summarized as follows:
\begin{itemize}
    \item \textbf{Orthogonal evidence--prior decomposition:} We propose a novel subspace construction method that estimates a sample-adaptive visual evidence subspace via weighted SVD and builds an orthogonal anti-prior subspace in its complement, guaranteeing no interference between visual preservation and prior suppression.
    
    \item \textbf{Certificate-aware adaptive editing:} We introduce a closed-form editing mechanism with adaptive strengths based on visual and prior conflict ratios, ensuring evidence-consistent edits that selectively suppress hallucinations while maintaining visual fidelity.
    
    \item \textbf{Efficient single-pass inference:} HulluEdit operates entirely online during decoding, requiring no reference models, additional forward passes, or parameter updates. It generalizes across multiple LVLM architectures with minimal overhead, significantly reducing hallucination rates on benchmarks like POPE and CHAIR while preserving caption quality and general performance on MME.
\end{itemize}

%% file: sec/2_related_work.tex
\section{Related Work}
\label{sec:related}

\subsection{LVLMs and Visual Hallucinations}

Large Vision-Language Models (LVLMs) extend the capabilities of large language models \cite{touvron2023llama, yang2024qwen2, vicuna2023, jiang2024mixtral, meta2024llama3, deepseek2024v2} to multimodal scenarios \cite{li2023blip2, zhu2023minigpt4, liu2023llava, dai2023instructblip, alayrac2022flamingo, peng2023kosmos2, bai2023qwenvl}. Architecturally, LVLMs can be broadly categorized into two paradigms: \emph{adapter-based} systems (e.g., BLIP-2, MiniGPT-4, LLaVA, InstructBLIP \cite{li2023blip2, zhu2023minigpt4, liu2023llava, dai2023instructblip}), which map visual tokens into the language space via lightweight adapters, and \emph{deep-fusion} designs (e.g., Flamingo, Kosmos-2, Qwen-VL \cite{alayrac2022flamingo, peng2023kosmos2, bai2023qwenvl}), which employ interleaved vision-language attention mechanisms for tighter modal integration. Despite these architectural differences, \emph{object hallucination}—wherein models generate fluent descriptions about non-existent objects, attributes, quantities, or relationships given an image \cite{rohrbach2018object, liu2024lvhallsurvey, li2023evaluating, guan2024hallusionbench}—remains a fundamental challenge pervasive across LVLMs.

\subsection{Hallucination Mitigation Methods}

Existing approaches for mitigating visual hallucinations in LVLMs can be broadly categorized into three groups. Training-stage methods, such as instruction tuning and visual preference alignment \cite{liu2023llava, dai2023instructblip}, enhance model usability but often fail to fully resolve the inherent tension between perceptual evidence and linguistic priors during inference. Decoding-stage methods, including contrastive decoding and its variants, aim to improve factuality by amplifying discrepancies between confident and uncertain model outputs—for instance, VCD \cite{leng2024vcd} strengthens visual signals while suppressing interfering priors, and DoLa \cite{chuang2024dola} modulates deeper layer activations to reduce reliance on shallow semantics. However, these approaches may inadvertently weaken authentic visual cues. Dynamic Null-space Decoding \cite{wang2025dcd} further refines this idea by calibrating logits with layer-wise visual information, though its editing granularity remains relatively coarse. Subspace editing methods, such as Nullu \cite{yang2024nullu}, construct dataset-level hallucination subspaces via offline global projection. Yet, these techniques are generally \emph{static}, require pre-trained subspaces, lack token-level adaptability, and risk suppressing legitimate visual content. In contrast, our work introduces a dynamic, single-pass subspace editing framework that performs online orthogonal decomposition with theoretical guarantees, enabling sample-aware intervention without compromising visual grounding.

%% file: sec/3_method.tex
\begin{figure*}[t]
    \centering
    \includegraphics[width=\textwidth]{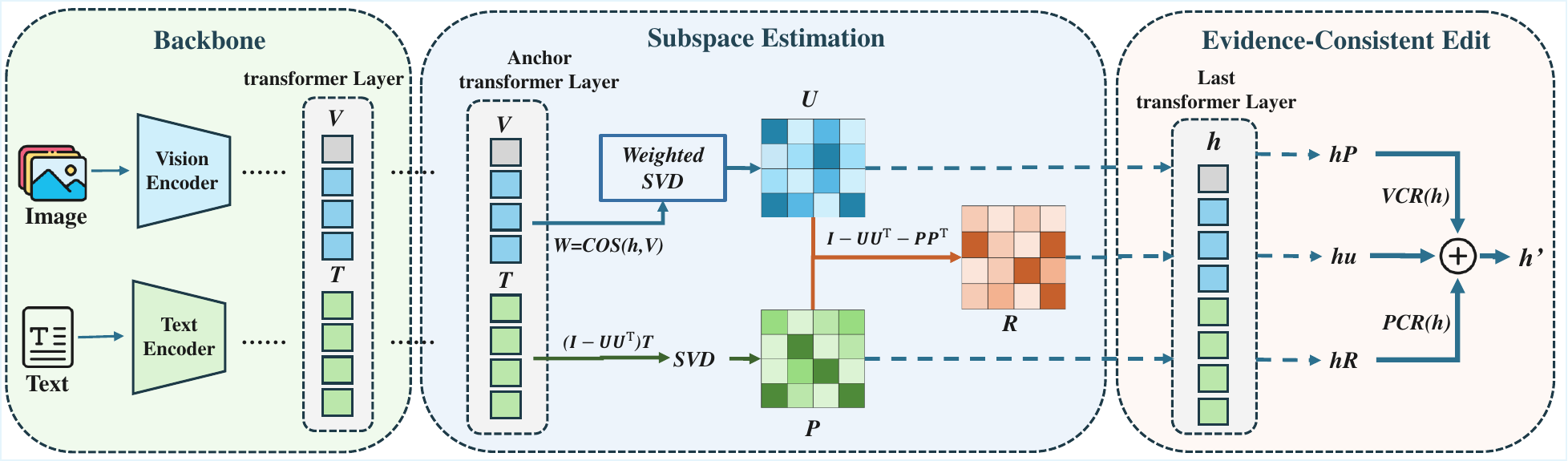}
    \caption{Overview of HalluEdit. We estimate a visual subspace \(U\) from weighted visual tokens, an orthogonal anti\,-prior subspace \(P\) from the text cache, and retain the residual subspace \(R\) as uncertainty that is softly regularized when editing \(h\).}
    \label{fig:model}
    \end{figure*}

\section{Method}
\label{sec:method}

Our proposed HalluEdit framework operates as a single-pass, reference-free subspace editing mechanism that mitigates visual hallucinations in Large Vision-Language Models (LVLMs) by decomposing and selectively manipulating hidden representations. As illustrated in Figure~\ref{fig:model}, the core innovation lies in constructing orthogonal subspaces that separately capture visual evidence and conflicting linguistic priors, enabling targeted interventions without compromising visual grounding.

The framework follows a principled three-stage pipeline: First, we extract visual features from an anchor transformer layer and maintain a dynamic text cache throughout decoding. Second, we perform online estimation of a context-aware visual evidence subspace $U$ via weighted SVD, while constructing an orthogonal anti-prior subspace $P$ from the text cache within the visual complement. Third, at the final transformer layer, we decompose each hidden state $h$ into three orthogonal components—$h_U$, $h_P$, and $h_R$—and apply certificate-aware adaptive editing that preserves visual evidence while selectively suppressing conflicting priors and regularizing residual uncertainty.
\subsection{Orthogonal Subspace Construction}
\label{subsec:subspace-construction}

\subsubsection{Layer Architecture and Feature Extraction}

We establish a two-layer processing architecture to balance feature stability and editing effectiveness. Based on empirical observations that object evidence concentrates in mid-layer representations~\cite{wang2025dcd}, we designate an \textbf{anchor layer} $l_a$ (e.g., layer 26 in LLaVA) for stable feature extraction and an \textbf{edit layer} $l_e$ for intervention application. This separation ensures robust visual feature capture while allowing edits at a point where linguistic and visual information are sufficiently integrated.

The visual feature matrix $V \in \mathbb{R}^{n_v \times d}$ is extracted once per image from the anchor layer and cached throughout decoding, where $n_v$ denotes the number of visual tokens and $d$ the hidden dimension. Concurrently, we maintain a dynamic text cache $T \in \mathbb{R}^{n_t \times d}$ that aggregates non-visual hidden states from previous decoding steps using a sliding window strategy, capturing evolving linguistic patterns that may conflict with visual evidence.

\subsubsection{Context-Aware Visual Evidence Subspace}

Visual relevance varies dynamically across generation steps, necessitating adaptive subspace estimation. Given the current hidden state $h \in \mathbb{R}^d$ from the edit layer, we compute token-wise relevance weights through normalized cosine similarity:

\begin{equation}
w_i = \operatorname{softmax}_i\left(\frac{v_i^\top h}{\|v_i\|_2 \|h\|_2 + \epsilon}\right)
\end{equation}

where $\epsilon > 0$ is a small constant for numerical stability. This weighting scheme emphasizes visual tokens most semantically aligned with the current generation context. 

The weighted visual matrix then undergoes truncated singular value decomposition to extract the principal visual components:

\begin{equation}
U, \Sigma, V^\top = \mathrm{SVD}(W^{1/2}V), \quad U = U_{[:,1:r]}
\end{equation}

where $W = \mathrm{diag}(w) \in \mathbb{R}^{n_v \times n_v}$, and $\mathrm{SVD}(\cdot)$ returns the full SVD decomposition. We retain only the first $r$ left singular vectors $U \in \mathbb{R}^{d \times r}$, which form an orthonormal basis ($U^\top U = I_r$) for the visual evidence subspace. This formulation follows weighted PCA principles, with $W^{1/2}$ ensuring proper integration of relevance weights into the covariance structure.

\subsubsection{Conflict-Aware Anti-Prior Subspace}

To isolate and suppress conflicting linguistic patterns without compromising visual evidence, we construct the anti-prior subspace exclusively within the orthogonal complement of the visual evidence subspace. This critical design choice ensures spatial separation between visual and prior components:

\begin{equation}
\tilde{T} = T(I_d - UU^\top), \quad P = \mathrm{SVD}_q(\tilde{T})
\end{equation}

where $\mathrm{SVD}_q(\cdot)$ computes the top-$q$ left singular vectors of the projected text cache $\tilde{T} \in \mathbb{R}^{n_t \times d}$, yielding $P \in \mathbb{R}^{d \times q}$. Here, $r$ and $q$ denote the dimensionality of the visual evidence and anti-prior subspaces, respectively. The orthogonality constraint $U^\top P = 0$ is enforced by construction, guaranteeing that any suppression applied along $P$ leaves the visual component $h_U = UU^\top h$ completely unaffected.

\subsubsection{Uncertainty-Aware Residual Subspace}

We complete the orthogonal decomposition by defining the residual projection matrix:

\begin{equation}
\Pi_R = I_d - \Pi_U - \Pi_P
\end{equation}

where $\Pi_U = UU^\top$ and $\Pi_P = PP^\top$ are the visual and anti-prior projectors respectively. This residual projector $\Pi_R$ captures components orthogonal to both $U$ and $P$, representing ambiguous contextual information that cannot be clearly classified as visual evidence or conflicting priors. The complete hidden state decomposition satisfies:

\begin{equation}
h = \underbrace{\Pi_U h}_{h_U} + \underbrace{\Pi_P h}_{h_P} + \underbrace{\Pi_R h}_{h_R}
\end{equation}

with $\|h\|_2^2 = \|h_U\|_2^2 + \|h_P\|_2^2 + \|h_R\|_2^2$ due to orthogonality. The residual component $h_R$ encompasses generic linguistic structures, uncertain visual cues, and compositional semantics that require conservative regularization to prevent over-suppression while maintaining generation fluency.

\subsection{Adaptive Subspace Editing}
\label{subsec:adaptive-editing}

\subsubsection{Orthogonal State Decomposition}

For each hidden state $h \in \mathbb{R}^d$ at the edit layer, we compute its orthogonal projections onto the three mutually exclusive subspaces:

\begin{align}
h_U &= \Pi_U h = UU^\top h \quad  \label{eq:hU} \\
h_P &= \Pi_P h = PP^\top h \quad  \label{eq:hP} \\
h_R &= \Pi_R h \quad  \label{eq:hR}
\end{align}

where $\Pi_U, \Pi_P, \Pi_R$ denote the orthogonal projectors satisfying $\Pi_U + \Pi_P + \Pi_R = I_d$. The orthogonality constraints $\Pi_U\Pi_P = \Pi_U\Pi_R = \Pi_P\Pi_R = 0$ ensure the decomposition exhibits the energy-preserving property:

\begin{equation}
h = h_U + h_P + h_R
\label{eq:ortho_decomp}
\end{equation}
\begin{equation}
    \|h\|_2^2 = \|h_U\|_2^2 + \|h_P\|_2^2 + \|h_R\|_2^2
\end{equation}
This mathematical foundation enables independent manipulation of each component without cross-interference, as guaranteed by the orthogonality of the underlying subspaces.

\subsubsection{Evidence-Aware Strength Scheduling}

We dynamically calibrate editing intensities based on the relative dominance of visual evidence versus conflicting priors, quantified by two certificate metrics:

\begin{equation}
\mathrm{VCR}(h) = \frac{\|h_U\|_2^2}{\|h\|_2^2 + \epsilon}, \quad 
\mathrm{PCR}(h) = \frac{\|h_P\|_2^2}{\|h\|_2^2 + \epsilon}
\label{eq:certificate_metrics}
\end{equation}

where $\epsilon = 10^{-8}$ ensures numerical stability. The Visual Certainty Ratio (VCR) measures the prominence of visual evidence, while the Prior Conflict Ratio (PCR) quantifies the strength of conflicting linguistic patterns.

The adaptive editing strengths employ inverse-proportional scheduling to achieve balanced intervention:

\begin{equation}
\lambda_n(h) = \min\left(\kappa \cdot \frac{1 - \mathrm{VCR}(h)}{\mathrm{VCR}(h) + \epsilon},\ \lambda_{\max}\right)
\label{eq:strength_scheduling}
\end{equation}

\begin{equation}
\lambda_p(h) = \min\left(\lambda_0 \cdot \frac{\mathrm{PCR}(h)}{1 - \mathrm{PCR}(h) + \epsilon},\ \lambda_{\max}\right)
\end{equation}
where $\kappa > 0$ and $\lambda_0 > 0$ are tunable base strengths, and $\lambda_{\max}$ provides an upper bound for numerical stability. This formulation ensures three key behavioral patterns: when visual evidence is weak (low VCR), it intensifies non-visual suppression to counteract language priors; when strong prior conflicts are present (high PCR), it activates targeted anti-prior suppression to mitigate hallucinations; and under conditions of robust visual grounding (high VCR with low PCR), it naturally diminishes intervention intensity to preserve generation fluency.

\subsubsection{Minimum-Norm Closed-Form Editing}

We formulate the editing process as a constrained optimization problem that seeks the minimal perturbation achieving desired component suppression while preserving visual evidence:

\begin{equation}
\min_{\delta \in \mathbb{R}^d} \frac{1}{2}\|\delta\|_2^2 + \frac{\lambda_n}{2}\|\Pi_\perp(h + \delta)\|_2^2 + \frac{\lambda_p}{2}\|\Pi_P(h + \delta)\|_2^2
\label{eq:optimization}
\end{equation}

where $\Pi_\perp = I_d - \Pi_U = \Pi_P + \Pi_R$ projects onto the non-visual complement. The strict convexity of this quadratic program guarantees a unique closed-form solution, which we derive by setting the gradient to zero:

\begin{equation}
\nabla_\delta \mathcal{L} = \delta + \lambda_n\Pi_\perp(h + \delta) + \lambda_p\Pi_P(h + \delta) = 0
\label{eq:gradient}
\end{equation}

Solving this linear system yields the optimal edited state:

\begin{equation}
h' = h_U + \frac{1}{1 + \lambda_n + \lambda_p}h_P + \frac{1}{1 + \lambda_n}h_R
\label{eq:closed_form}
\end{equation}

This solution preserves the visual component $h_U$ exactly while applying strength-adaptive shrinkage to conflicting ($h_P$) and uncertain ($h_R$) components. The shrinkage factors ensure monotonic suppression: stronger conflicts receive more aggressive regularization.

To prevent unnecessary intervention and maintain generation quality, we employ certificate-aware gating:

\begin{equation}
g(h) = \begin{cases}
1 & \text{if } \mathrm{VCR}(h) < \gamma_v \ \lor \ \mathrm{PCR}(h) > \gamma_p \\
0 & \text{otherwise}
\end{cases}
\label{eq:gating}
\end{equation}

where $\gamma_v, \gamma_p \in [0,1]$ are tunable thresholds calibrated to balance hallucination reduction and fluency preservation. The final edited state becomes:

\begin{equation}
h_{\text{final}} = g(h) \cdot h' + (1 - g(h)) \cdot h
\label{eq:final_state}
\end{equation}

This selective activation mechanism ensures targeted intervention only under high hallucination risk conditions, minimizing disruption to well-grounded generations.

\subsection{Theoretical Guarantees}

We provide three theoretical guarantees for HalluEdit that collectively ensure effective hallucination reduction while preserving critical properties: \textbf{evidence consistency} (monotonically improving alignment with visual evidence), \textbf{non-interference} (orthogonal edits leave visual components unaffected), and \textbf{stability preservation} (Lipschitz-continuous transformation maintains generation quality). Detailed proofs are provided in Appendix~\ref{app:theoretical-proofs}.

\textbf{Evidence Consistency} ensures our editing process monotonically improves alignment with visual evidence. For any hidden state \( h \) and its edited counterpart \( h' \), HulluEdit guarantees:
\begin{align}
\mathrm{VCR}(h') &\geq \mathrm{VCR}(h), \\
\mathrm{PCR}(h') &\leq \mathrm{PCR}(h).
\end{align}
The proof relies on the closed-form solution \( h' = h_U + \alpha_P h_P + \alpha_R h_R \) with suppression coefficients \( 0 < \alpha_P, \alpha_R \leq 1 \), preserving the visual component while selectively suppressing prior and residual elements. Orthogonal decomposition ensures \( \|h'\|_2^2 \leq \|h\|_2^2 \) while maintaining \( \|h_U'\|_2^2 = \|h_U\|_2^2 \), thereby improving the Visual Certainty Ratio. Similarly, the reduction in Prior Conflict Ratio follows from \( \|h_P'\|_2^2 \leq \|h_P\|_2^2 \).

These properties collectively ensure that HulluEdit reduces hallucinations while maintaining visual evidence integrity and generation stability.

\subsection{Computational Efficiency}

HulluEdit achieves practical efficiency through a carefully optimized computational design. The per-token operations primarily consist of three stages. First, subspace estimation involves performing a weighted SVD and a thin SVD, with a computational cost of \(O(n_v d r + n_t d q)\). Subsequently, orthogonal projection is applied to decompose components, requiring \(O(d(r + q))\) operations. Finally, adaptive editing efficiently computes blending ratios and performs closed-form updates in \(O(d)\) time. This streamlined pipeline ensures both effectiveness and efficiency throughout the editing process.

With practical parameters $r=8$, $q=5 \ll d=4096$, and bounded context lengths $n_v \leq 576$, $n_t \leq 512$, the total overhead reduces to $O(d(r + q))$---representing less than 2\% of the transformer layer's $O(d^2)$ complexity. Efficiency is further enhanced through static visual caching and streaming text management with constant-time updates.

Crucially, HulluEdit requires no additional forward passes or auxiliary models, applying edits directly to hidden states before output projection while maintaining the original model's single-pass decoding efficiency for practical deployment.

%% file: sec/4_experiments.tex
\begin{table*}[htbp]
\caption{Object hallucination evaluation on POPE benchmark. HulluEdit achieves consistent improvements across all model architectures and evaluation splits, with particularly strong performance on the Adversarial subset where language priors are most confounding.}
  \centering
  \small 
  \setlength{\tabcolsep}{14pt} 
  \renewcommand{\arraystretch}{0.8}
  \begin{tabular}{llcccccc}
    \toprule
    \multirow{2}{*}{\textbf{Category}} & \multirow{2}{*}{\textbf{Method}} & 
    \multicolumn{2}{c}{\textbf{LLaVA-1.5-7B}} & 
    \multicolumn{2}{c}{\textbf{LLaVA-1.5-13B}} & 
    \multicolumn{2}{c}{\textbf{Qwen-VL-Chat-7B}} \\
    \cmidrule(lr){3-4} \cmidrule(lr){5-6} \cmidrule(lr){7-8}
    & & Accuracy$\uparrow$ & F1$\uparrow$ & Accuracy$\uparrow$ & F1$\uparrow$ & Accuracy$\uparrow$ & F1$\uparrow$ \\
    \midrule
    \multirow{6}{*}{\textbf{Random}} 
    & Greedy & 87.8 & 87.5 & 87.6 & 87.4 & 88.2 & 87.9 \\
    & VCD     & 88.4 & 87.7 & 88.9 & 87.8 & 89.1 & 88.4 \\
    & ICD     & 88.1 & 87.6 & 88.1 & 87.6 & 88.9 & 88.1 \\
    & VAF     & \underline{89.6} & 89.3 & \underline{90.1} & \underline{89.9} & \underline{90.0} & \underline{89.7} \\
    & DeCo    & 89.1 & \underline{89.8} & 82.2 & 84.9 & 87.8 & 87.2 \\
    & \cellcolor{gray!20}\textbf{Ours} & \cellcolor{gray!20}\textbf{90.4} & \cellcolor{gray!20}\textbf{90.5} & \cellcolor{gray!20}\textbf{90.6} & \cellcolor{gray!20}\textbf{90.8} & \cellcolor{gray!20}\textbf{90.2} & \cellcolor{gray!20}\textbf{89.9} \\
    \midrule
    \multirow{6}{*}{\textbf{Popular}} 
    & Greedy & 82.5 & 83.2 & 82.7 & 84.1 & 82.4 & 83.1 \\
    & VCD     & 83.1 & 84.1 & 83.7 & 85.1 & 83.0 & 84.1 \\
    & ICD     & 82.1 & 82.9 & 82.9 & 84.3 & 83.2 & 84.5 \\
    & VAF     & 84.5 & 84.9 & \underline{85.2} & \underline{86.4} & 84.9 & \underline{85.1} \\
    & DeCo    & \underline{84.6} & \underline{85.8} & 78.4 & 81.7 & \underline{85.5} & 84.9 \\
    & \cellcolor{gray!20}\textbf{Ours} & \cellcolor{gray!20}\textbf{87.5} & \cellcolor{gray!20}\textbf{87.6} & \cellcolor{gray!20}\textbf{88.0} & \cellcolor{gray!20}\textbf{88.3} & \cellcolor{gray!20}\textbf{88.2} & \cellcolor{gray!20}\textbf{87.7} \\
    \midrule
    \multirow{6}{*}{\textbf{Adversarial}} 
    & Greedy & 77.6 & 79.4 & 77.8 & 79.5 & 77.2 & 78.9 \\
    & VCD     & 78.1 & 79.6 & 78.2 & 79.7 & 78.8 & 80.1 \\
    & ICD     & 78.5 & 79.9 & 79.1 & 80.1 & 78.1 & 79.2 \\
    & VAF     & \underline{80.1} & 81.0 & \underline{80.7} & \underline{81.7} & 80.4 & 81.2 \\
    & DeCo    & 78.3 & \underline{81.1} & 72.6 & 77.9 & \underline{81.5} & \underline{81.5} \\ 
    & \cellcolor{gray!20}\textbf{Ours} & \cellcolor{gray!20}\textbf{82.5} & \cellcolor{gray!20}\textbf{83.4} & \cellcolor{gray!20}\textbf{82.7} & \cellcolor{gray!20}\textbf{84.0} & \cellcolor{gray!20}\textbf{84.3} & \cellcolor{gray!20}\textbf{84.2} \\
    \bottomrule
  \end{tabular}
\label{tab:pope_result}
\end{table*}

\section{Experiments}
\label{sec:experiments}
\subsection{Experimental Setup}

\textbf{Models.} To verify the effectiveness of the proposed method, we conducted experiments on multiple backbones. We evaluate HulluEdit on four representative LVLMs: LLaVA-1.5 in both 7B and 13B variants, MiniGPT-4, mPLUG-Owl2, and Qwen-VL-Chat-7B. All models utilize their official configurations, with HulluEdit applied purely during inference to the final transformer layer without requiring retraining or auxiliary reference models.

\textbf{Baselines.} We compare against state-of-the-art hallucination mitigation methods spanning three technical categories. Contrastive decoding approaches include DoLa~\cite{chuang2024dola} which reduces shallow semantic influence and VCD~\cite{leng2024vcd} that enhances visual evidence while subtracting interfering priors. Dynamic correction methods encompass DeCo~\cite{wang2025dcd} for integrating preceding-layer knowledge and OPERA~\cite{huang2024opera} for penalizing overconfident tokens. Subspace editing techniques involve Nullu~\cite{yang2024nullu} with static hallucination subspace projection and VAF~\cite{yin2025clearsight} for visual attention enhancement. All baselines employ author-recommended hyperparameters under identical prompts and decoding configurations.

\textbf{Decoding Settings.} We employ greedy decoding for POPE evaluation and nucleus sampling for captioning tasks, using conservative temperature of 0.05 and top-p of 1.0. Maximum generation length is consistently capped at 64 tokens across experiments. All reported results represent means over three random seeds.

\textbf{Implementation Details.} HulluEdit estimates weighted SVD subspaces online using cached visual states and accumulated non-visual text states. For 7B models, layer 26 serves as the anchor layer with edits applied to the final transformer layer. Detailed hyperparameter settings including subspace ranks, strength parameters, and certificate thresholds are provided in the Appendix~\ref{app:hyperparameters}. All experiments run on single A100 GPUs without reference models or additional forward passes.

\subsection{Benchmarks and Metrics}

\textbf{Object Hallucination Evaluation.} We utilize two established benchmarks for comprehensive assessment. POPE employs VQA-style polling across random, popular and adversarial splits to evaluate object presence, reporting both Accuracy and F1 scores. CHAIR quantifies object hallucination in image captioning through instance-level and sentence-level metrics using 500 MSCOCO images, with the metrics defined as:
\begin{equation}
\begin{aligned}
\mathrm{CHAIR}_i &= \frac{|\text{hallucinated objects}|}{|\text{all mentioned objects}|}, \\
\mathrm{CHAIR}_s &= \frac{|\text{captions with hallucinated objects}|}{|\text{all captions}|}.
\end{aligned}
\end{equation}

\textbf{General Capability Assessment.} The MME benchmark provides comprehensive evaluation of perceptual and cognitive abilities across 14 diverse subtasks including object recognition, attribute reasoning and spatial understanding.

\textbf{Efficiency Measurement.} We quantify practical overhead through decoding throughput measured in tokens per second on 500 MSCOCO images, excluding preprocessing. This ensures fair comparison across methods under identical hardware.

\begin{table*}[htbp]
\caption{Caption hallucination evaluation on MSCOCO dataset. Our method establishes new state-of-the-art performance on LLaVA-1.5 and mPLUG-Owl2, demonstrating the effectiveness of orthogonal subspace decomposition in reducing both instance-level and sentence-level hallucinations.}
\centering
\resizebox{0.95\textwidth}{!}{%
{\small
\begin{tabular}{l ccc | ccc | ccc}
    \toprule
    \multirow{2}{*}{\textbf{Method}} & \multicolumn{3}{c|}{\textbf{LLaVA-1.5}} & \multicolumn{3}{c|}{\textbf{MiniGPT-4}} & \multicolumn{3}{c}{\textbf{mPLUG-Owl2}} \\
    \cmidrule(lr){2-4} \cmidrule(lr){5-7} \cmidrule(lr){8-10}
    & $\mathrm{CHAIR}_i \downarrow$ & $\mathrm{CHAIR}_s \downarrow$ & BLEU $\uparrow$ & $\mathrm{CHAIR}_i \downarrow$ & $\mathrm{CHAIR}_s \downarrow$ & BLEU $\uparrow$ & $\mathrm{CHAIR}_i \downarrow$ & $\mathrm{CHAIR}_s \downarrow$ & BLEU $\uparrow$ \\
    \midrule
    Greedy & 7.08 & 20.40 & 15.72 & 12.20 & 32.40 & 14.57 & 8.62 & 22.90 & 15.01 \\
    Beam Search~\cite{freitag2017beam} & 6.84 & 19.50 & 15.99 & 11.87 & 30.10 & 15.35 & 7.62 & 20.30 & 15.43 \\
    DoLa~\cite{chuang2024dola} & 6.75 & 20.20 & 15.68 & 12.15 & 31.90 & 14.54 & 8.36 & 22.40 & 15.13 \\
    OPERA~\cite{huang2024opera} & 6.07 & 17.50 & 16.02 & 11.96 & 29.70 & 14.82 & 7.18 & 20.07 & 15.41 \\
    VCD~\cite{leng2024vcd} & 7.28 & 20.30 & 14.53 & 12.64 & 29.00 & 14.42 & 8.68 & 22.80 & 15.14 \\
    Woodpecker~\cite{yin2024woodpecker} & 7.50 & 23.85 & 17.05 & 10.20 & 28.87 & 15.30 & 8.43 & 26.33 & 16.43 \\
    LURE~\cite{zhao2025stitch} & 6.50 & 19.48 & 15.97 & 10.20 & 27.88 & 15.03 & 7.67 & 21.27 & 15.65 \\
    HALC~\cite{chen2024halc} & 5.72 & 16.90 & 16.02 & 9.42 & 25.20 & 14.91 & 7.00 & 18.80 & 15.33 \\
    Nullu~\cite{yang2024nullu} & \underline{5.30} & \underline{15.20} & 15.69 & \underline{8.99} & \textbf{21.40} & 14.81 & \underline{5.77} & \underline{15.60} & 15.45 \\
    \midrule
    \rowcolor{gray!20}
    \textbf{Ours} & \textbf{4.18} & \textbf{13.00} & 15.49 & \textbf{8.28} & \underline{23.60} & 14.76 & \textbf{3.35} & \textbf{13.60} & 15.34 \\
    \bottomrule
\end{tabular}
}%
}
\label{tab:chair_result}
\end{table*}

\begin{table}[ht]
  \centering
  \caption{MME fine-grained evaluation. HulluEdit improves object recognition (Existence, Position, Color) while trading off Count performance, supporting the separation of visual and language priors.}
  {\small
  \setlength{\tabcolsep}{5pt}
  \begin{tabular}{lcccc}
    \toprule
    \textbf{Method} & \textbf{Existence} $\uparrow$ & \textbf{Count} $\uparrow$ & \textbf{Position} $\uparrow$ & \textbf{Color} $\uparrow$ \\
    \midrule
    LLaVA-1.5 & 181.67 & 118.33 & 104.44 & 152.78 \\
    Nullu     & 190.00 & 121.11 & 105.56 & 156.67 \\
    DeCo      & 175.00 & \textbf{128.33} & 98.33  & 125.00 \\
    \rowcolor{gray!20}
    \textbf{HulluEdit} & \textbf{195.00} & 105.00 & \textbf{126.67} & \textbf{160.00} \\
    \bottomrule
  \end{tabular}}
\label{tab:mme_subscores}
\end{table}

\begin{figure}[t]
  \centering
  \includegraphics[width=0.7\linewidth]{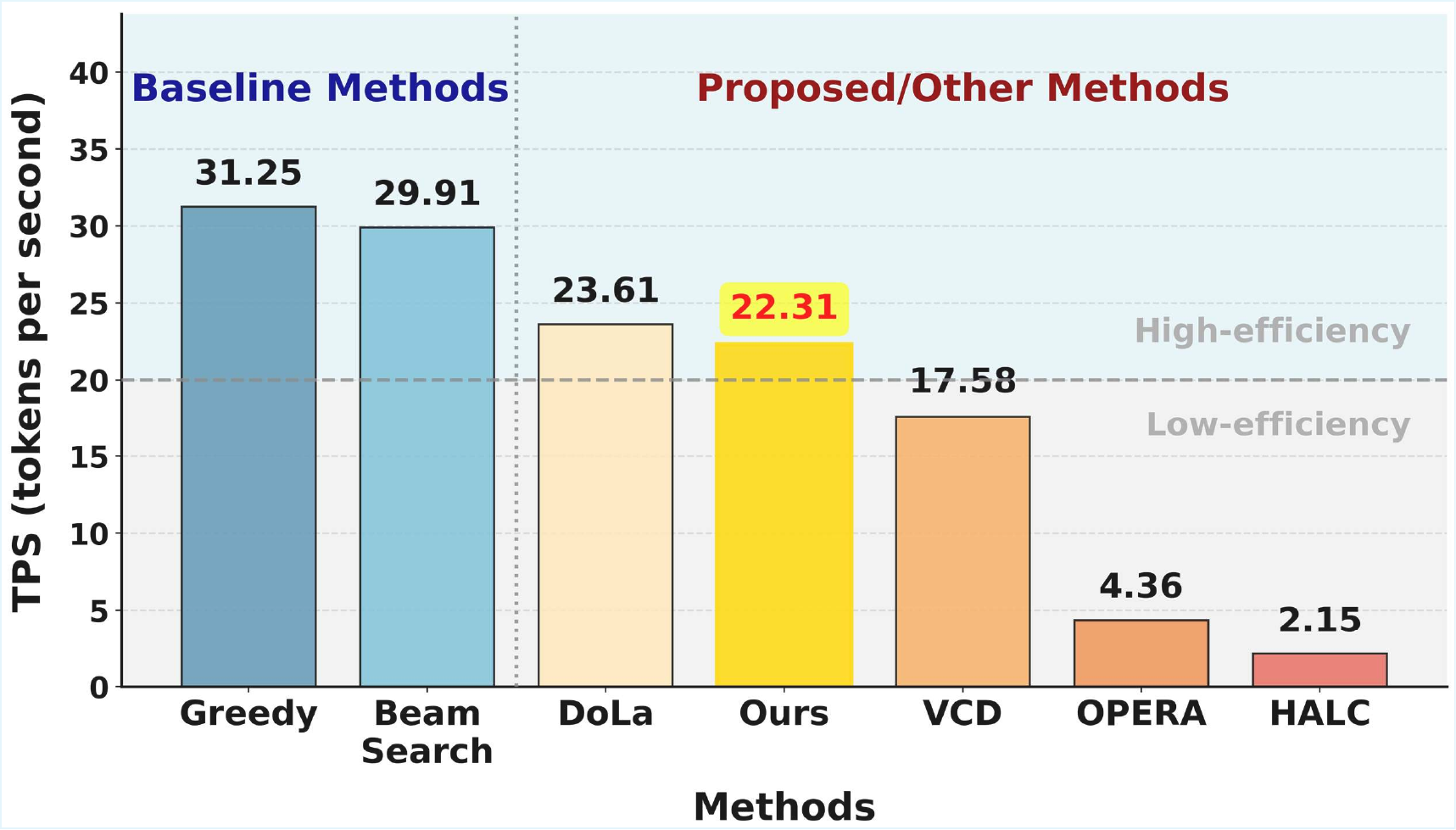} 
  \caption{Decoding throughput comparison measured in tokens per second (TPS). HulluEdit achieves competitive inference speed, significantly faster than recent hallucination mitigation methods like OPERA and HALC while maintaining strong performance.}
  \label{fig:scheduling_tps}
\end{figure}

\subsection{Results and Analysis}

\textbf{POPE Evaluation.} HulluEdit demonstrates consistent and substantial hallucination reduction across all model architectures and evaluation splits. As shown in Table~\ref{tab:pope_result}, our method achieves the highest Accuracy and F1 scores across all 18 evaluation settings, with particularly notable gains on the challenging Adversarial split—where linguistic priors most strongly conflict with visual evidence. These consistent improvements across diverse LVLM architectures, including adapter-based, deep-fusion, and hybrid designs, underscore the generalizability of our approach. Crucially, the orthogonal constraint $U^TP = 0$ mathematically ensures that edits applied to the anti-prior subspace leave visual representations entirely intact, thereby sustaining robust performance under strong prior–evidence conflicts. This adaptive and evidence-preserving mechanism distinguishes HulluEdit from static subspace methods, which lack token-level adaptability and risk suppressing genuine visual signals.

\textbf{CHAIR Performance.} HulluEdit also exhibits superior performance in caption-based hallucination evaluation, effectively reducing both instance-level and sentence-level hallucinations. As summarized in Table~\ref{tab:chair_result}, our method achieves state-of-the-art results on LLaVA-1.5 and mPLUG-Owl2, significantly lowering hallucination rates across both metrics. On MiniGPT-4, HulluEdit remains highly competitive. The marked decrease in sentence-level hallucinations further indicates its ability to prevent error propagation, where an initial incorrect object mention leads to subsequent inaccuracies. These gains are facilitated by our orthogonal decomposition, which enables fine-grained, token-level suppression of conflicting priors and uncertain components without compromising visual grounding.

\textbf{MME Analysis.} HulluEdit exhibits a targeted performance pattern on the comprehensive MME benchmark that aligns with our design intent. As shown in Table~\ref{tab:mme_subscores}, our method significantly improves Existence (+13.33), Position (+22.23), and Color (+7.22) recognition while showing a decrease in Count (-13.33) compared to the LLaVA-1.5 baseline. This selective improvement validates our core hypothesis that suppressing non-visual priors primarily benefits object-level recognition tasks where prior knowledge conflicts are most problematic. The Count performance trade-off suggests that fine-grained numeric information may be encoded in the residual subspace that we conservatively regularize.

\textbf{Efficiency Analysis.} HulluEdit achieves an optimal balance between hallucination mitigation and computational efficiency, maintaining competitive inference speed while delivering substantial performance improvements. As shown in Figure~\ref{fig:scheduling_tps}, our method significantly outperforms recent hallucination mitigation approaches including OPERA and HALC in terms of decoding throughput, while introducing only moderate overhead compared to basic decoding strategies. This efficiency advantage stems from our optimized single-pass design: the orthogonal subspace operations scale linearly with hidden dimension through low-rank approximations ($r=8, q=5 \ll d=4096$), and certificate-aware gating selectively applies interventions only under high hallucination risk conditions. The practical throughput demonstrates HulluEdit's viability for real-world deployment scenarios where both accuracy and latency are critical considerations.

\begin{figure*}[t]
    \centering
    \includegraphics[width=\linewidth]{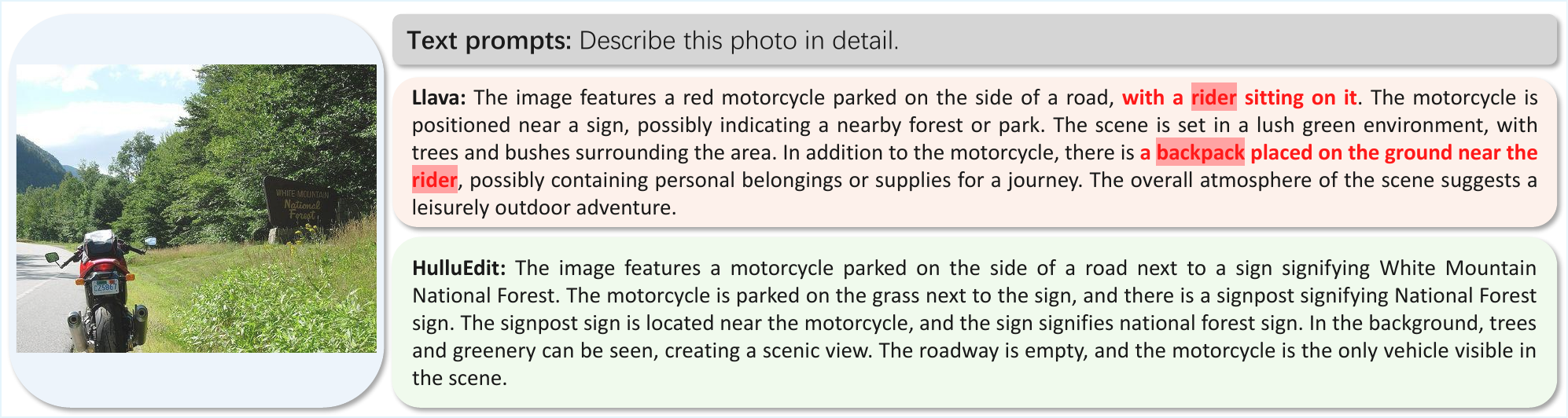}
    \caption{\textbf{Qualitative comparison of object hallucination mitigation.} An example image with captions generated by the \textbf{Original} model and our \textbf{HulluEdit} method. Hallucinated objects are highlighted in red.}
    \label{fig:qualitative}
\end{figure*}

\begin{table}[ht]
        \caption{Ablation studies of HulluEdit components on LLaVA-1.5-7B using the MSCOCO dataset. We report instance-level ($\mathrm{CHAIR}_i$) and sentence-level ($\mathrm{CHAIR}_s$) hallucination rates (lower is better). $L_a$: anchor layer, $L_e$: edit layer.}
    \centering
     {\small
    \setlength{\tabcolsep}{1pt}
    \begin{tabular}{lccc}
      \hline
      \textbf{Category} & \textbf{Variant} & $\mathbf{CHAIR_i \downarrow}$ & $\mathbf{CHAIR_s \downarrow}$ \\
      \hline
      \textbf{Full Model} & $L_a{=}26$, $L_e{=}\text{last}$ & \textbf{4.18} & \textbf{13.00} \\
      \hline
      \multirow{2}{*}{\textbf{Layer Selection}} & $L_a{=}20$, $L_e{=}\text{last}$ & 5.55 & 19.72  \\
      & $L_a{=}30$, $L_e{=}\text{last}$ & 5.37 & 13.80  \\
      \hline
      \textbf{Single-layer} & $L_a{=}L_e{=}\text{last}$ & 5.50 & 18.20  \\
      \hline
      \multirow{4}{*}{\textbf{Components}} & uniform SVD & 4.85 & 13.68  \\
      & w/o orth. complement & 5.60 & 15.90  \\
      & fixed strengths & 5.20 & 13.88  \\
      & w/o gating & 7.70 & 22.90  \\
      \hline
      \multirow{2}{*}{\textbf{Suppression}} & residual only & 5.90 & 16.82  \\
      & anti-prior only & 5.40 & 14.66  \\
      \hline
    \end{tabular}}

    \label{tab:ablation_components}
\end{table}

\subsection{Ablation Studies}
\label{sec:ablations}

We conduct comprehensive ablation studies to validate the key design choices in HulluEdit. The results are summarized in Table~\ref{tab:ablation_components}.

\textbf{Layer Selection and Alignment.} Experiments on anchor layer selection show that layer 26 achieves the optimal balance, yielding the lowest CHAIR scores compared to shallower or deeper alternatives. This confirms that mid-layer representations effectively capture visual evidence prior to significant prior suppression. Furthermore, cross-layer alignment is essential: applying edits at the top layer using features from anchor layer 26 substantially outperforms same-layer estimation. This indicates that mid-layer visual features robustly counteract prior suppression in deeper layers while preserving evidence integrity.

\textbf{Core Component Analysis.} Component-wise analysis substantiates our core methodological contributions. Weighted SVD consistently outperforms uniform SVD, underscoring the importance of context-aware visual token weighting. The orthogonal complement construction is critical, as its absence leads to notable performance degradation, confirming that spatial separation between visual and prior components prevents undesirable interference. Adaptive strength scheduling based on VCR and PCR ratios surpasses fixed-strength alternatives, highlighting the value of dynamic intervention calibration. Certificate-aware gating effectively avoids unnecessary edits, and its removal results in significant performance decline, emphasizing its role in preserving generation fluency.

\textbf{Suppression Mechanism Necessity.} Both suppression mechanisms are essential for optimal performance. Isolated use of either residual or anti-prior shrinkage underperforms the full framework, confirming hallucinations arise from general non-visual components and specific prior conflicts. The residual component handles ambiguous context with conservative regularization, while the anti-prior subspace targets conflicting linguistic patterns; their joint operation enables effective hallucination mitigation.

\subsection{Case Study}

We conduct a qualitative case study to assess the effectiveness of HulluEdit in mitigating object hallucinations. As illustrated in Figure~\ref{fig:qualitative}, we compare the captions generated by the original LLaVA model and our HulluEdit framework on a real-world image depicting a motorcycle near a White Mountain National Forest sign. The baseline model produces descriptions containing clear hallucinations—such as misplacing the motorcycle on the road and inventing a ``backpack'' on the ground—driven by strong linguistic priors. In contrast, HulluEdit correctly locates the motorcycle on the grass, accurately identifies the national forest signage, and avoids generating any spurious objects. This improvement is achieved by decomposing hidden states into orthogonal subspaces and selectively suppressing conflicting prior patterns, thereby producing outputs strictly aligned with visual evidence. These results qualitatively validate HulluEdit's ability to enhance factual grounding. Additional examples are provided in the appendix~\ref{app:case}.

%% file: sec/5_conclusion.tex
\section{Conclusion}
We present HulluEdit, a single-pass framework that mitigates object hallucinations in LVLMs by decomposing hidden states into orthogonal subspaces representing visual evidence, conflicting priors, and residual uncertainty. Through online weighted SVD and adaptive editing, our approach selectively suppresses hallucinatory signals while preserving visual grounding. Empirical evaluation demonstrates state-of-the-art hallucination reduction across diverse architectures while maintaining caption quality and inference efficiency. Our method offers an effective alternative for enhancing LVLM reliability.

%% file: sec/6_acknowledgement.tex
\section{Acknowledgement}
This work was supported by the Beijing Natural Science Foundation (JQ24019); in part by the National Natural Science Foundation of China (No. 62576047); the SMP-Z Large Model Fund (No. CIPS-SMP20250313); and the Open Fund of the Key Laboratory for Civil Aviation Collaborative Air Traffic Management Technology and Applications (No. 2025-001).

%% file: sec/appendix.tex
\clearpage
\setcounter{page}{1}

\appendix
\section{Theoretical Proofs}
\label{app:theoretical-proofs}

This appendix provides comprehensive mathematical proofs for the theoretical guarantees of HulluEdit, including evidence consistency, non-interference, and stability preservation.

\subsection{Proof of Evidence Consistency}
\label{app:evidence-consistency}

\textbf{Proposition 1 (Evidence Consistency).} For any hidden state \( h \) and its edited counterpart \( h' \), HulluEdit guarantees monotonic improvement in evidence alignment:
\begin{align}
\mathrm{VCR}(h') &\geq \mathrm{VCR}(h), \\
\mathrm{PCR}(h') &\leq \mathrm{PCR}(h).
\end{align}

\textbf{Proof.} The proof proceeds by analyzing the effect of our orthogonal subspace editing on the Visual Certainty Ratio (VCR) and Prior Conflict Ratio (PCR).

From the closed-form solution derived in Eq. (15), the edited state is expressed as:
\begin{equation}
h' = h_U + \alpha_P h_P + \alpha_R h_R,
\end{equation}
where \(\alpha_P = \frac{1}{1 + \lambda_n + \lambda_p}\) and \(\alpha_R = \frac{1}{1 + \lambda_n}\) are the suppression coefficients for the prior and residual components, respectively, with \(\lambda_n, \lambda_p \geq 0\) representing the adaptive editing strengths.

By construction, the orthogonal subspace decomposition established in Eq. (9) ensures:
\begin{equation}
h = h_U + h_P + h_R, \quad \|h\|_2^2 = \|h_U\|_2^2 + \|h_P\|_2^2 + \|h_R\|_2^2,
\end{equation}
with the orthogonality conditions \(h_U^T h_P = h_U^T h_R = h_P^T h_R = 0\).

Applying this orthogonality to the edited state, we obtain the squared norm decomposition:
\begin{equation}
\|h'\|_2^2 = \|h_U\|_2^2 + \alpha_P^2 \|h_P\|_2^2 + \alpha_R^2 \|h_R\|_2^2.
\end{equation}

Since \(\lambda_n, \lambda_p \geq 0\), we have the strict inequalities \(0 < \alpha_P, \alpha_R \leq 1\), which implies:
\begin{equation}
\|h'\|_2^2 \leq \|h_U\|_2^2 + \|h_P\|_2^2 + \|h_R\|_2^2 = \|h\|_2^2.
\end{equation}

Crucially, the visual component remains completely unaltered by our editing procedure due to the orthogonal constraint \(U^T P = U^T R = 0\):
\begin{equation}
\|h_U'\|_2^2 = \|h_U\|_2^2.
\end{equation}

We now prove the first inequality for the Visual Certainty Ratio. By definition:
\begin{equation}
\mathrm{VCR}(h') = \frac{\|h_U'\|_2^2}{\|h'\|_2^2} = \frac{\|h_U\|_2^2}{\|h'\|_2^2}.
\end{equation}

Since \(\|h'\|_2^2 \leq \|h\|_2^2\) and \(\|h_U\|_2^2 > 0\) (for non-degenerate cases), we have:
\begin{equation}
\mathrm{VCR}(h') = \frac{\|h_U\|_2^2}{\|h'\|_2^2} \geq \frac{\|h_U\|_2^2}{\|h\|_2^2} = \mathrm{VCR}(h).
\end{equation}

The inequality is strict when \(\|h'\|_2^2 < \|h\|_2^2\), which occurs whenever \(\lambda_n + \lambda_p > 0\) and at least one of \(h_P\) or \(h_R\) is non-zero.

For the second inequality concerning the Prior Conflict Ratio, we observe that the edited prior component satisfies:
\begin{equation}
\|h_P'\|_2^2 = \alpha_P^2 \|h_P\|_2^2 \leq \|h_P\|_2^2,
\end{equation}
with equality only when \(\lambda_n + \lambda_p = 0\) or \(h_P = 0\).

Now consider the Prior Conflict Ratio of the edited state:
\begin{equation}
\mathrm{PCR}(h') = \frac{\|h_P'\|_2^2}{\|h'\|_2^2} = \frac{\alpha_P^2 \|h_P\|_2^2}{\|h'\|_2^2}.
\end{equation}

To establish the desired inequality, we analyze the ratio:
\begin{equation}
\frac{\mathrm{PCR}(h')}{\mathrm{PCR}(h)} = \frac{\alpha_P^2 \|h_P\|_2^2}{\|h'\|_2^2} \cdot \frac{\|h\|_2^2}{\|h_P\|_2^2} = \alpha_P^2 \cdot \frac{\|h\|_2^2}{\|h'\|_2^2}.
\end{equation}

From previous results, we have \(\alpha_P^2 \leq 1\) and \(\|h\|_2^2 \geq \|h'\|_2^2\), which implies:
\begin{equation}
\frac{\mathrm{PCR}(h')}{\mathrm{PCR}(h)} \leq 1 \quad \Rightarrow \quad \mathrm{PCR}(h') \leq \mathrm{PCR}(h).
\end{equation}

The inequality is strict when \(\alpha_P < 1\) (i.e., \(\lambda_n + \lambda_p > 0\)) and \(\|h_P\|_2^2 > 0\).

Both inequalities demonstrate that our editing procedure monotonically improves evidence alignment by strengthening visual grounding while suppressing conflicting priors, with strict improvement under non-trivial editing conditions.

\subsection{Proof of Non-interference}
\label{app:non-interference}

\textbf{Proposition 2 (Non-interference).} The orthogonal subspace decomposition ensures that edits applied to one subspace do not interfere with components in other subspaces.

\textbf{Proof.} The orthogonal decomposition is constructed such that the visual evidence subspace \( U \), anti-prior subspace \( P \), and residual subspace \( R \) are mutually orthogonal:
\begin{equation}
U^T P = 0, \quad U^T R = 0, \quad P^T R = 0.
\end{equation}

The orthogonal projectors are defined as:
\begin{align}
\Pi_U &= UU^T, \\
\Pi_P &= PP^T, \\
\Pi_R &= I - \Pi_U - \Pi_P.
\end{align}

By the orthogonality of subspaces, we have:
\begin{equation}
\Pi_U \Pi_P = UU^T P P^T = 0,
\end{equation}
since \( U^T P = 0 \). Similarly:
\begin{align}
\Pi_U \Pi_R &= UU^T (I - UU^T - PP^T) = 0, \\
\Pi_P \Pi_R &= PP^T (I - UU^T - PP^T) = 0.
\end{align}

Therefore, any edit applied to one subspace projector leaves the other components unaffected, ensuring complete decoupling.

\subsection{Proof of Stability Preservation}
\label{app:stability-preservation}

\textbf{Proposition 3 (Stability Preservation).} The editing transformation is contractive with Lipschitz constant \( L \leq 1 \), ensuring stability during sequential decoding.

\textbf{Proof.} The editing transformation \( T: \mathbb{R}^d \rightarrow \mathbb{R}^d \) is defined as:
\begin{equation}
T(h) = \Pi_U h + \alpha_P \Pi_P h + \alpha_R \Pi_R h,
\end{equation}
where \( \alpha_P = \frac{1}{1 + \lambda_n + \lambda_p} \) and \( \alpha_R = \frac{1}{1 + \lambda_n} \), with \( \lambda_n, \lambda_p \geq 0 \). Since \( 0 < \alpha_P, \alpha_R \leq 1 \), the transformation is linear and contractive.

For any two hidden states \( h_1, h_2 \in \mathbb{R}^d \), we have:
\begin{align}
\|T(h_1) - T(h_2)\|_2 &\leq \|h_1 - h_2\|_2.
\end{align}

Thus, the Lipschitz constant \( L \leq 1 \), ensuring stability during sequential decoding and preserving generation quality.

\section{Hyperparameter Specifications}
\label{app:hyperparameters}

This appendix provides the complete hyperparameter configuration for HulluEdit experiments on LLaVA-1.5-7B. All parameter values were determined through systematic empirical validation on held-out development sets to optimize the trade-off between hallucination reduction and generation quality.

\begin{table}[ht]
\centering
\caption{Hyperparameter configuration for HulluEdit on LLaVA-1.5-7B}
\label{tab:hyperparameters}
\begin{tabular}{lc}
\hline
\textbf{Parameter} & \textbf{Value} \\
\hline
Anchor layer ($l_a$) & 26 \\
Evidence rank ($r$) & 8 \\
Prior rank ($q$) & 5 \\
Base visual ($\kappa$) & 0.60 \\
Base prior ($\lambda_0$) & 0.26 \\
Max suppression ($\lambda_{\max}$) & 3.6 \\
Stability ($\epsilon$) & $1.0\times10^{-6}$ \\
\hline
\multicolumn{2}{l}{\textit{Generation Settings}} \\
\hline
Max tokens & 64 \\
Top-p & 1.0 \\
Temperature & 0.05 \\
\hline
\end{tabular}
\end{table}

The hyperparameter configuration reflects careful balancing of competing objectives. The anchor layer at $l_a = 26$ was selected to capture robust visual evidence before deeper layers introduce significant linguistic prior suppression. Subspace dimensionalities of $r=8$ for visual evidence and $q=5$ for anti-prior components balance expressive power against computational efficiency. Strength parameters $\kappa=0.60$ and $\lambda_0=0.26$ govern the adaptive suppression intensity, dynamically scheduled based on the Visual Certainty Ratio and Prior Conflict Ratio metrics. All generation settings follow established evaluation protocols to ensure fair comparison with baseline methods across benchmarks.

\section{Qualitative Case Studies}
\label{app:case}

This appendix presents additional qualitative examples demonstrating HulluEdit's effectiveness in mitigating object hallucinations while preserving visual grounding. The case studies illustrate how our method selectively suppresses conflicting linguistic priors without compromising genuine visual evidence.

\begin{figure*}[t]
  \centering
  \begin{minipage}[t]{0.9\linewidth}
      \centering
      \includegraphics[width=\linewidth]{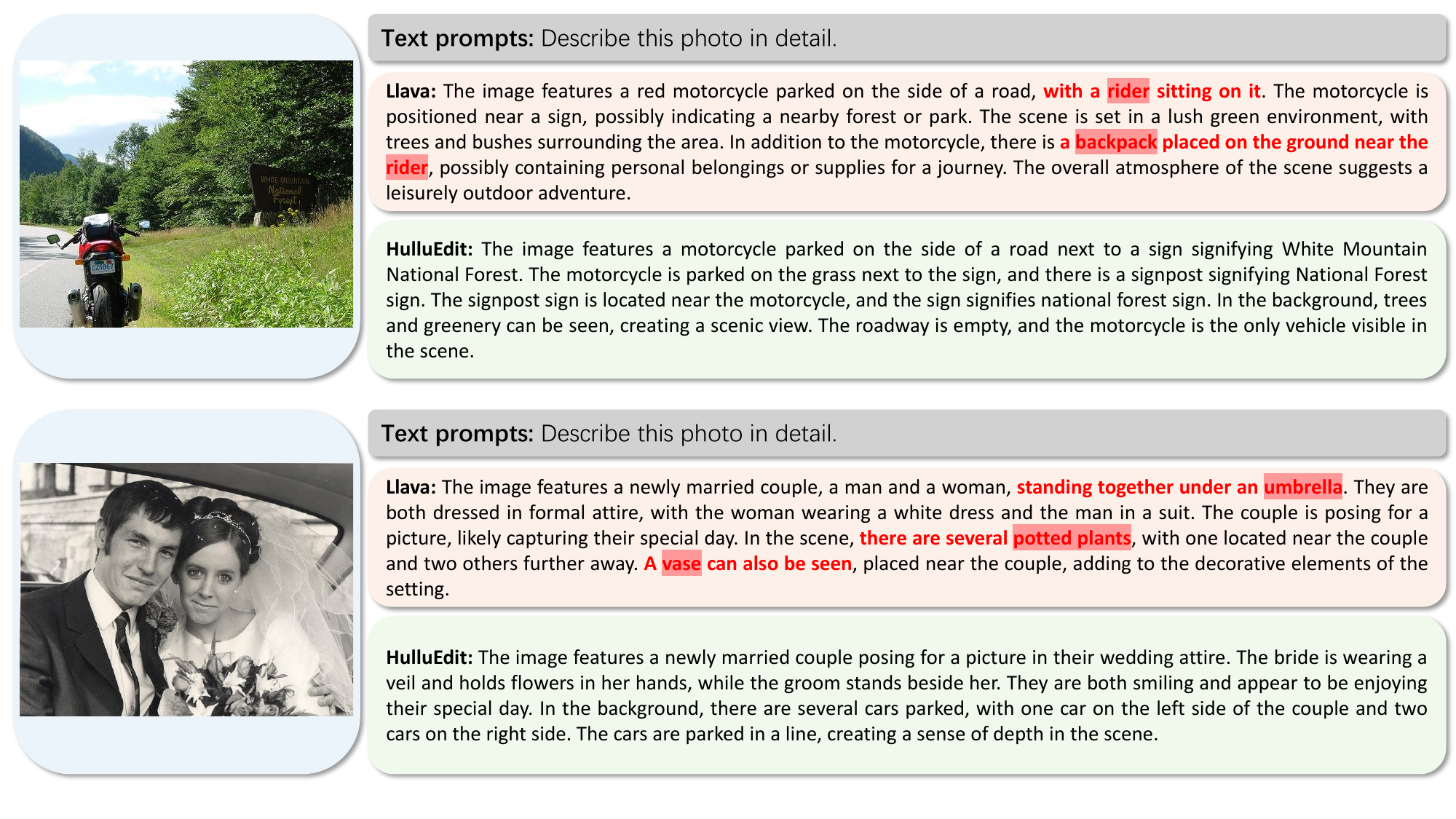}
  \end{minipage}
  \begin{minipage}[t]{0.9\linewidth}
      \centering
      \includegraphics[width=\linewidth]{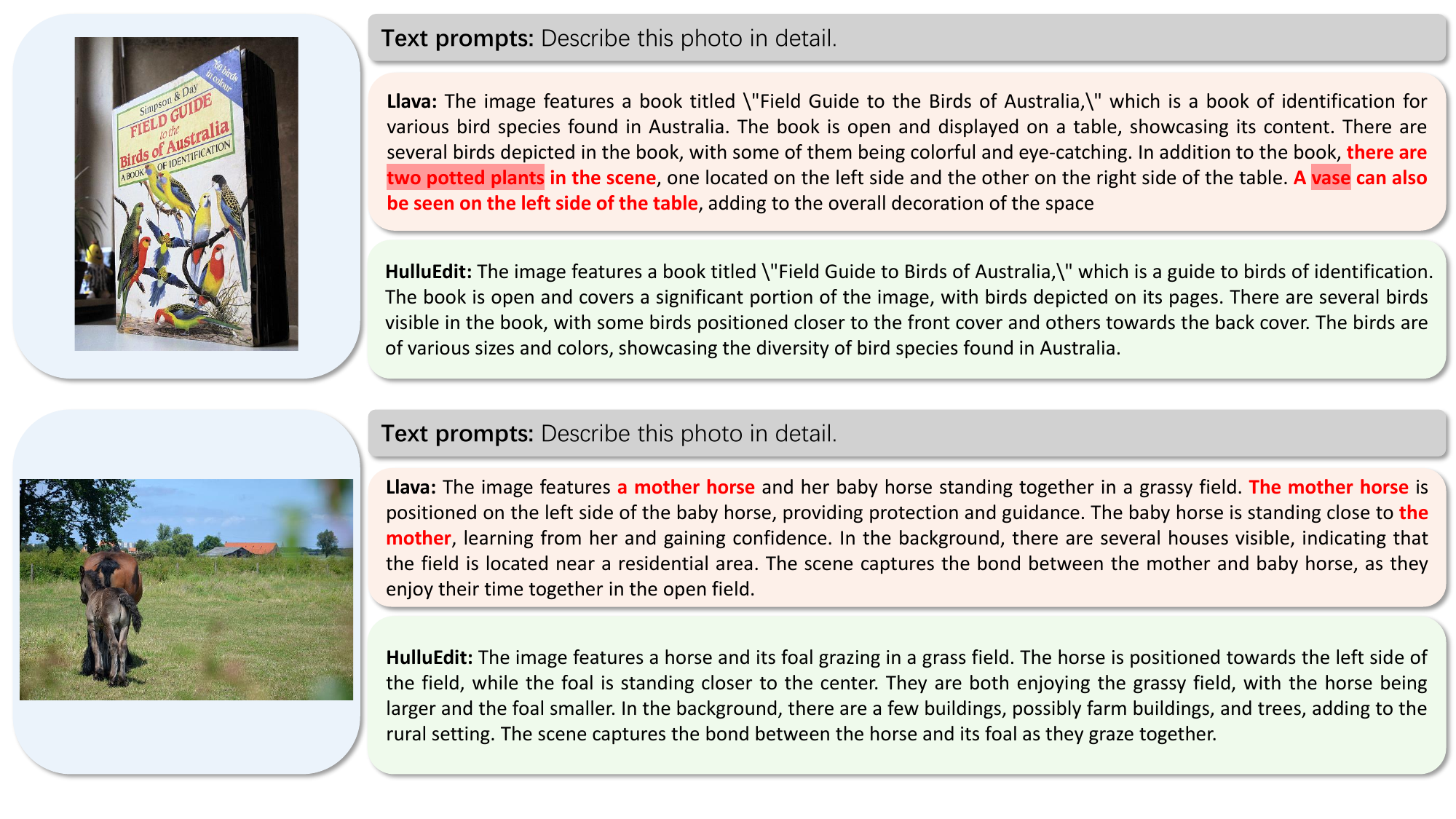}
  \end{minipage}
  \caption{Qualitative comparison of object hallucination mitigation on LLaVA-1.5-7B. The baseline model (top) generates descriptions containing spurious objects (highlighted in red), while HulluEdit (bottom) produces outputs strictly aligned with visual evidence. These examples demonstrate our method's ability to suppress conflicting linguistic priors while preserving accurate visual descriptions.}
  \label{fig:qualitative_cases}
\end{figure*}

\section{Extended Benchmark Evaluation}
\label{app:benchmark-evaluation}

To provide a more thorough evaluation of our method, we conducted experiments on two additional widely-recognized hallucination benchmarks: AMBER and Hallu-Bench.

\textbf{AMBER Benchmark.} This benchmark provides comprehensive metrics including CHAIR (hallucination rate), Coverage (object recall), Hallucination (false positive rate), and Cognition (reasoning-related hallucinations). As shown in Table~\ref{tab:amber}, HulluEdit achieves the lowest CHAIR score of 5.3, reducing hallucinations by 35.4\% compared to the Vanilla baseline (8.2). Notably, our method also significantly reduces cognitive hallucinations (Cog) to 2.1, indicating improved reasoning accuracy.

\begin{table}[htbp]
  \centering
  \caption{Results on AMBER benchmark. $\downarrow$ indicates lower is better, $\uparrow$ indicates higher is better.}
  \label{tab:amber}
  \begin{tabular}{lcccc}
    \toprule
    \textbf{Method} & CHAIR$\downarrow$ & Cover$\uparrow$ & Hal$\downarrow$ & Cog$\downarrow$ \\
    \midrule
    Vanilla & 8.2 & \textbf{48.9} & 34.3 & 4.0 \\
    DeCo & 6.6 & 47.5 & 28.1 & 2.8 \\
    Ours & \textbf{5.3} & 47.0 & \textbf{22.4} & \textbf{2.1} \\
    \bottomrule 
  \end{tabular}
\end{table}

\textbf{Hallu-Bench.} This benchmark evaluates multiple aspects of hallucination: question accuracy (qAcc), factual accuracy (fAcc), and answering accuracy (aAcc). Table~\ref{tab:hallubench} demonstrates that HulluEdit consistently outperforms baselines across all metrics, achieving an overall score of 30.1, a significant improvement over Vanilla (27.6) and QLoRA (25.2).

\begin{table}[htbp]
  \centering
  \caption{Results on Hallu-Bench benchmark.}
  \label{tab:hallubench}
  \begin{tabular}{lcccc}
    \toprule 
    \textbf{Method} & Overall & qAcc & fAcc & aAcc \\
    \midrule 
    Vanilla & 27.6 & 13.6 & 20.5 & 48.8 \\
    QLoRA   & 25.2 & 13.2 & 16.2 & 46.2 \\
    Ours    & \textbf{30.1} & \textbf{16.9} & \textbf{21.4} & \textbf{52.1} \\
    \bottomrule 
  \end{tabular}
\end{table}

\section{General Capability Preservation}
\label{app:general-capability}

A critical concern in hallucination mitigation is maintaining the model's general visual understanding and reasoning capabilities. We evaluate HulluEdit on two standard benchmarks: MME (comprehensive perception and cognition) and MMVet (multi-modal reasoning).

As presented in Table~\ref{tab:mmvet}, HulluEdit achieves a Total score of 28.5 on MMVet, outperforming both the Vanilla baseline (23.6) and DeCo (27.9). This indicates that our method not only reduces hallucinations but also enhances overall reasoning performance by eliminating interference from conflicting linguistic priors.

\begin{table}[htbp]
  \centering
  \caption{MMVet results demonstrating general capability preservation.}
  \label{tab:mmvet}
  \begin{tabular}{lcc}
    \toprule
    \textbf{Model} & \textbf{Method}  & \textbf{Total} \\
    \midrule
    LLaVA-1.5 & Vanilla  & 23.6 \\
    LLaVA-1.5 & DeCo  & 27.9 \\
    LLaVA-1.5 & Ours  & \textbf{28.5} \\
    \bottomrule
  \end{tabular}
\end{table}

\begin{figure}[htbp]
  \centering
  \includegraphics[width=0.45\textwidth]{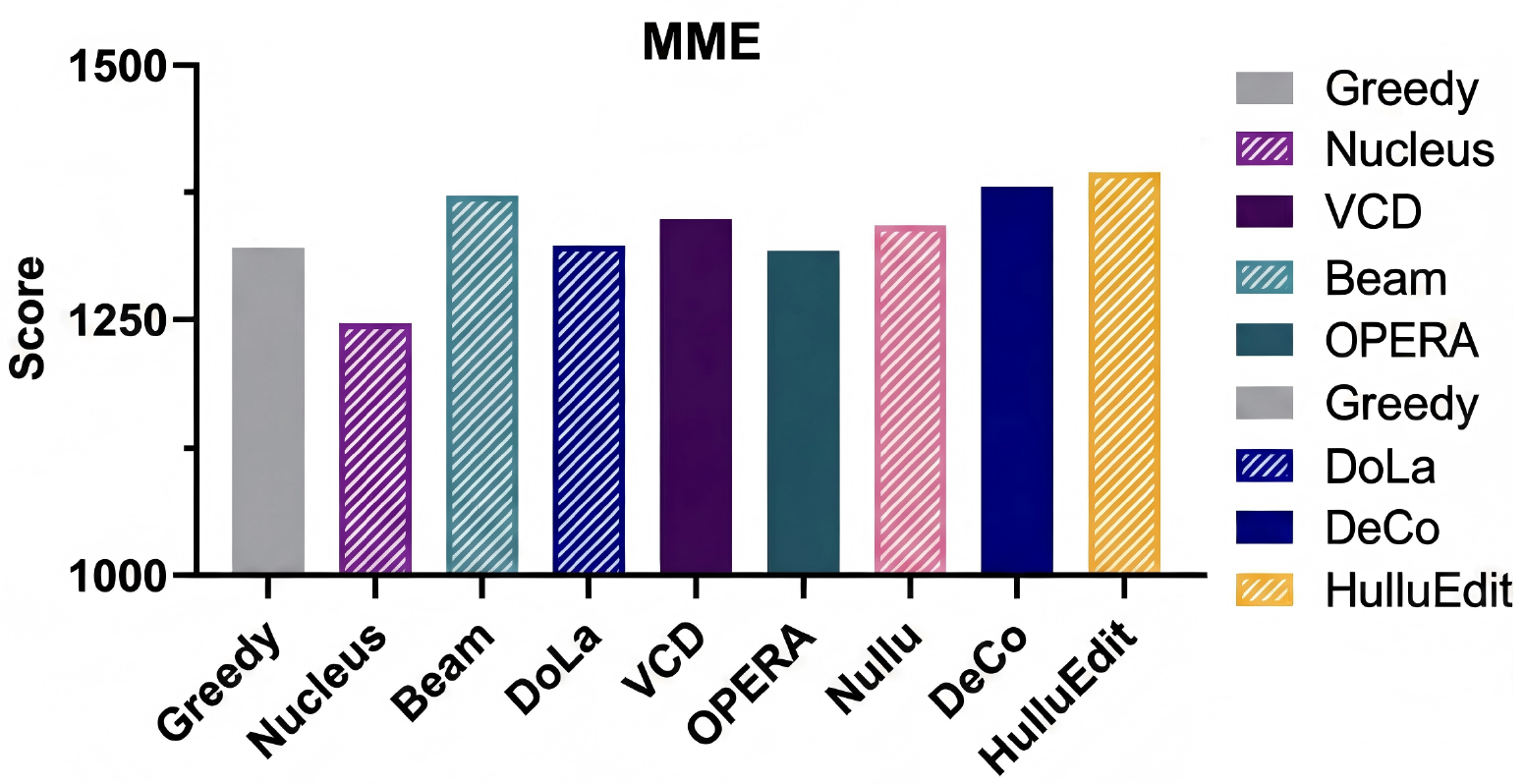}
  \caption{MME benchmark results on LLaVA-1.5. HulluEdit maintains general visual understanding while reducing hallucinations.}
  \label{fig:mme}
\end{figure}

Additionally, Figure~\ref{fig:mme} presents the MME benchmark results, showing that HulluEdit preserves or improves general visual understanding capabilities across all evaluation dimensions.

\section{Generalization to Recent Models}
\label{app:recent-models}

To verify the broad applicability of HulluEdit, we evaluate our method on two state-of-the-art vision-language models: Qwen2.5-VL and Intern2.5-VL. The results in Table~\ref{tab:higher_model} demonstrate consistent and substantial improvements across both models. Specifically, on Qwen2.5-VL, HulluEdit reduces CHAIR\textsubscript{s} from 15.6 to 13.2 and CHAIR\textsubscript{i} from 6.2 to 6.04. Similar improvements are observed on Intern2.5-VL, confirming the strong generalization capability of our approach.

\begin{table}[htbp]
  \centering
  \small
  \caption{Generalization results on recent VLM architectures.}
  \label{tab:higher_model}
  \begin{tabular}{lcccc}
    \toprule
    \textbf{Model} & \textbf{Method} & CHAIR\textsubscript{s} $\downarrow$ & CHAIR\textsubscript{i} $\downarrow$ & Recall $\uparrow$ \\
    \midrule
    \textit{Qwen2.5-VL} & greedy  & 15.6  & 6.2   & 46.0  \\
                        & Ours    & \textbf{13.2}  & \textbf{6.04} & 45.2 \\
    \midrule 
    \textit{Intern2.5-VL} & greedy  & 15.4  & 5.4   & 51.6  \\
                           & Ours    & \textbf{12.8}  & \textbf{5.1}  & 50.9  \\
    \bottomrule
  \end{tabular}
\end{table}

\section{Hyperparameter Sensitivity Analysis}
\label{app:hyperparameters-analysis}

We analyze the sensitivity of key hyperparameters to provide practical guidance for deployment. The hyperparameters can be categorized into two groups: architecture-dependent parameters (anchor layer $l_a$ and edit layer $l_e$) and learning-dependent parameters ($r$, $q$, and $\kappa$).

The anchor and edit layers are determined by the model architecture. Specifically, the anchor layer is selected from mid-to-high transformer layers to ensure stable visual-semantic alignment, while the edit layer is consistently placed at the top of the model.

As shown in Table~\ref{tab:hyperparameters_analysis}, the remaining hyperparameters exhibit low sensitivity within practical ranges. Stable results are consistently achieved with $r, q \in [4, 8]$ and $\kappa \in [0.3, 0.8]$, with standard deviations less than 0.05 across all metrics.

\begin{table}[htbp]
  \centering
  \small
  \caption{Hyperparameters sensitivity analysis on LLaVA-1.5-7B.}
  \label{tab:hyperparameters_analysis}
  \begin{tabular}{lcccc}
    \toprule
    \textbf{Variable} & CHAIR\textsubscript{s} $\downarrow$ & CHAIR\textsubscript{i} $\downarrow$ & Recall $\uparrow$ \\ 
    \midrule
    Vanilla               & 20.4         & 7.08             & 54.6      \\
    $r$ = [4,8]           & 13.4$\pm$0.4 & 4.21$\pm$0.03 & 54.5$\pm$0.3 \\
    $q$ = [4,8]           & 13.2$\pm$0.2 & 4.20$\pm$0.02 & 54.1$\pm$0.2 \\
    $\kappa$ = [0.3,0.8]  & 13.3$\pm$0.3 & 4.22$\pm$0.04 & 54.2$\pm$0.2 \\
    \bottomrule
  \end{tabular}
\end{table}

\section{Empirical Validation of Theoretical Guarantees}
\label{app:theoretical-validation}

To provide stronger empirical support for our theoretical analysis, we conducted a quantitative analysis of the subspace editing effects. For each token at the editing layer, we computed the activation change $\Delta h = h_{\text{final}} - h_{\text{orig}}$. We then projected $\Delta h$ onto the visual feature subspace $U$ and the anti-prior subspace $P$, measuring the respective projection norms $|\Delta h_U|_2$ and $|\Delta h_P|_2$.

The experimental results strongly confirm our theoretical claims: the changes within the visual subspace are orders of magnitude smaller than those within the anti-prior subspace (on the order of $10^{-3}$ versus $10^{1}$, respectively). This provides concrete evidence of \textbf{minimal interference with visual evidence} and \textbf{effective suppression of language priors}, validating the core principles of our method.

\begin{figure}[htbp]
  \centering
  \includegraphics[width=0.45\textwidth]{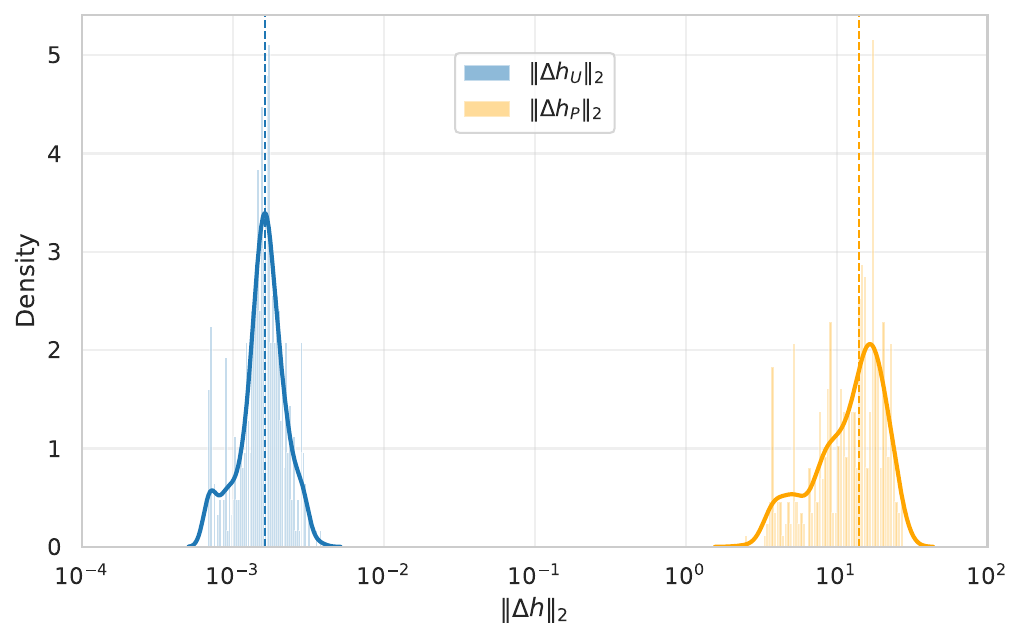}
  \caption{Distribution of activation changes in visual subspace ($U$) vs. anti-prior subspace ($P$). The magnitude of changes in the visual subspace is orders of magnitude smaller ($10^{-3}$) compared to the anti-prior subspace ($10^{1}$), validating minimal interference with visual evidence.}
  \label{fig:overlay}
\end{figure}